%% file: Template_REV1.tex
\documentclass{article}
\usepackage{spconf,amsmath,graphicx,amsfonts,mathrsfs,color}

\renewcommand{\eqref}[1]{Eq.~(\ref{#1})}
\title{Demystifying Deep Learning: A Geometric Approach to Iterative Projections}
\name{Ashkan~Panahi$^\dagger$, Hamid~Krim$^\dagger$, Liyi~Dai$^\dagger$\textsuperscript{\#}\sthanks{This work is partially supported by the U.S. Army Research Office under agreement W911NF-16-2-0005}}
\address{$^\dagger$ Department of Electrical and Computer Engineering, North Carolina State University,\\ Raleigh, NC, 27606\\
\# The U.S. Army Research Office, Durham, NC 27709}

\begin{document}
\input{fonts}
%
\maketitle
\begin{abstract}
Parametric approaches to Learning, such as deep learning (DL), are highly popular in nonlinear regression, in spite of their extremely difficult training with  their increasing complexity (e.g. number of layers in DL). In this paper, we present an alternative semi-parametric framework which foregoes the ordinarily required feedback, by introducing the novel idea of geometric regularization. We show that certain deep learning techniques such as  residual network (ResNet) architecture are closely related to our approach. Hence, our technique can be used to analyze these types of deep learning. Moreover, we present preliminary results which confirm that our approach can be easily trained to obtain complex structures.       
\end{abstract}
\section{Introduction}
\label{sec:intro}
Learning a nonlinear function through a finite number of input-output observations is a fundamental problem of supervised machine learning, and has wide applications in science and engineering. From a statistical vantage point, this problem entails a regression procedure which, depending on the nature of the underlying function, may be linear or nonlinear.  In the past few decades, there has been a flurry of advances in the area of nonlinear regression \cite{hastie2009overview}. Deep learning is perhaps one of the most well-known approaches with a promising and remarkable performance in great many applications.

Deep learning has a number of distinctive advantages: 1. It relies on a parametric description of functions that are easily computable. Once the parameters (weights) of a deep network are set, the output can be rapidly computed in a feed forward fashion by a few iterations of affine and elementwise nonlinear operations; 2.  it can avoid over-parametrization by adjusting the architecture (number of parameters) of the network, hence providing control over the generalization power of deep learning. Finally, deep networks have been observed to be highly flexible in expressing complex and highly nonlinear functions \cite{haykin2001neural,hassoun1995fundamentals}. There are, however,  a number of challenges associated with deep learning, chief among them is that of obtaining the exact assessment of their expressive power which remains to this day, an open problem.  An important exception is the single-layer network for which the so called universal approximation property (UAP) has been established for some time\cite{gybenko1989approximation, hornik1991approximation}, and which is clearly a highly desirable property. Another practical difficulty with deep learning is that the output becomes unproportionally sensitive to the parameters of different layers, making it, from an optimization perspective, extremely difficult to train \cite{lecun2015deep}. A recent solution is the so-called residual network (ResNet) learning, which introduces bridging branches to the conventional deep learning architecture \cite{he2016deep}. 
In this paper, we address the above issues by proposing a different perspective on learning with a substantially different architecture, which totally forgoes any feedback. Specifically, we propose an interative foward projection in lieu of back propagation to update parameters.  As such, this may  rapidly yield an over-parametrized system, we restrict each layer to perform an "incremental" update on the data, as approximately captured by the realization of a differential equation, we refer to  as  geometric regularization, as discussed in Section \ref{sec:GR}. The formulation of this geometric regularization allows us to tie the analysis of deep networks to differential geometry. The study in \cite{hauser2017principles} notices this relation, but adopts a different approach. In particular, we conjecture a converse of the celebrated Frobenius integrability theorem, which potentially proves a universal approximation property for a family of modified deep ResNets. We also present preliminary results in Section \ref{sec:results}, and show that foregoing back propagation in a neural network does not greatly limit the expressive power of deep networks, and in fact  potentially decreases their training effort, dramatically.    

\section{MMSE Estimation by Geometric Regularization}
For the sake of generality, we consider a $C^1$ Banach manifold\footnote{A Banach manifold is an infinite-dimensional generalization of a conventional differentiable manifold \cite{lang1972differential}.} $\calF$ of functions $f:\bbR^n\to\bbR^m$, where $n,m$ are the dimensions of the data and label vectors, respectively, and each element $f\in\calF$ represents a candidate model between the data and the labels. The arbitrary choice of $\calF$ allows one to impose structural properties on the models. Due to space limitation and for clarity sake,  we just focus on the simpler case of $\calF=\calL^2$, i.e. the space of square integrable functions,  and defer further generalizations to a later publication. Moreover, consider a probability space $(\Omega,\Sigma,\mu)$, and two random vectors $\bx:\Omega\to\bbR^n$ and $\by:\Omega\to\bbR^m$ representing statistical information about the data. As samples $(\bx_t,\by_t)$ for $t=1,2,\ldots,T$ of $\bx,\by$ are often provided, in which case their empirical distribution is used.
%

We consider the supervised learning problem by minimizing the following mean square error (MSE), 
\begin{equation}\label{eq:objective}
L(f)=\bbE\left[\|f(\bx)-\by\|_2^2\right],
\end{equation}
where $\bbE[\ldotp]$ denotes expectation. For observed samples $(\bx_t,\by_t)$, this criterion simplifies to
\begin{eqnarray}\label{eq:min_sample}
\minl_{f\in\calF}\frac 1 T\suml_{t=1}^T\|f(\bx_t)-\by_t\|_2^2.
\end{eqnarray}
In practice, the statistical assumptions in \eqref{eq:min_sample} are highly underdetermined and minimization of MSE (MMSE) leads to undesired solutions. To cope with this, additional constraints are considered to tame the problem by way of regularization. For example the set $\calF$ can be restricted to a (finite-dimensional) smooth sub-manifold. This is an implicit  fact  in parametric approaches, such as deep neural networks. 
\subsection{Geometric Regularization}\label{sec:GR}
We introduce a more general type of regularization, which also includes parametric restriction. Our generalization is inspired by the observation that standard (smooth) optimization techniques such as gradient descent, are based on a realization of a differential equation of the following form,
\begin{equation}\label{eq:integral}
\frac{\td f_\tau}{\td \tau}=\phi(f_\tau),
\end{equation}
where $\phi(f)\in T_f$ is a tangent vector of $\calF$ at $f$. The resulting solution is typically in an iterative form, as follows,
\begin{equation}\label{eq:step_simple}
f_{t+1}=f_t+\mu_t\phi(f_t),
\end{equation}
where $\mu_t$ is the step size at iteration $t=0,1,2,\ldots$. The tangent vector $\phi(f_\tau)$ is often selected as a descent direction, where according to \eqref{eq:objective}, $\td L/\td \tau<0$. 

For geometric regularization, we restrict the choice of the tangent vector to a closed cone $C_f\subseteq T_f$ in the tangent space. In the case of function estimation, where $\calF$ and hence the tangent space $T_f$, is infinite dimensional, we adopt a parametric definition of $C_f$ by restricting  the tangent vector to a finite dimensional space. However, this might not restrict the function to a finite dimensional submanifold. A particularly important case, where geometric regularization simplifies to a parametric (finite dimensional) manifold restriction is given by the Frobenius integrability theorem \cite{camacho2013geometric,khalil1996noninear}:

\begin{theorem}\label{theorem:fro}{\bf (Frobenius theorem)}
Suppose that $C_f$ is an  $n$-dimensional linear subspace of $T_f$. For any choice of $\phi(f)\in C_f$, the solution of \eqref{eq:integral} remains on an $n$-dimensional submanifold of $\calF$ only, depending on the initial point $f_0$, iff $C_f$ is involutive, i.e. for any two vector fields $\phi(f),\psi(f)$ in $C_f$ we have that
\[
\left[\phi(f),\psi(f)\right]\in C_f,
\]  
where $[\ldotp,\ldotp]$ denotes a Lie bracket \cite{khalil1996noninear}.
\end{theorem}
A simple example of an involutive regularization is when
\[
C_f=\left\{W_0f+\suml_{k=1}^rW_kf^k+b\mid W_k\in\bbR^{m\times m}, b\in\bbR^{ m}\right\}
\]
where $f^k$ are fixed functions. It is clear that the solution $f_t$ remains in $C_{f_0}$  from an initial $f_0$. Hence, this case corresponds to a linear regression. Selecting a nonlinear function $g:\bbR\to\bbR$, we can write a more general form of the geometric regularization discussed here, as follows,
\begin{eqnarray}\label{eq:GR_general}
&C_f=\nwl
&\left\{\Gamma(f)\left[W_0f+\suml_{k=1}^rW_kf^k+b\right]\mid W_k\in\bbR^{d\times d_k}, b\in\bbR^d\right\},
\end{eqnarray}
where $f^k$ are arbitrary fixed functions and $\Gamma(a)$ for $a=(a_1,a_2,\ldots,a_d)\in\bbR^d$ is a diagonal matrix with diagonals $\Gamma_{ii}=\td g/\td x(a_i)$. 

\section{Algorithmic Solution}
The solution to the differential equation in \eqref{eq:integral} with the geometric regularization in \eqref{eq:GR_general} requires a specification of the tangent vectors $\phi(f)\in C_f$. To preserve a good control on the computations, and much like for the DNN architecture, we define $f:\bbR^n\to\bbR^d$, where the reduced dimension $d<n$ is a design parameter. Then, the desired function is calculated as $Df_\tau+c$ where $D\in\bbR^{m\times d}$ and $c\in\bbR^m$ are fixed. Letting $f_0(x)=Ux$ where $U\in\bbR^{d\times n}$ is a constant dimensionality reduction matrix, we rewrite the MSE objective in \eqref{eq:objective} as
\[
L(f)=\bbE\left[\left\|\by-Df(\bx)-c\right\|_2^2\right].
\]
We subsequently apply the steepest descent principle to yield the following optimization:
\begin{eqnarray}\label{eq:steep}
\phi_f=\arg\minl_{\phi\in C_f\mid \|\phi\|_2\leq 1}\frac{\td L}{\td\tau}.
\end{eqnarray}
We next observe that under mild assumptions,
\[
\frac{\td L}{\td\tau}=-\bbE\left[\left\langle \bz~,~D\Gamma(f)\left[W_0f+\suml_{k=1}^rW_kf^k+b\right]\right\rangle\right],
\]
where $\bz=\by-Df(\bx)-c$ and $W_0,W_k,b$ are to be decided based on the optimization in \eqref{eq:steep}. After some manipulations, this leads to
\[
\phi_f=\Gamma(f)\left[W_{0,f}f+\suml_{k=1}^rW_{k,f}f^k+b_f\right],
\]
where
\begin{eqnarray}\label{eq:params}
&W_{0,f}=\bbE\left[\Gamma(f(\bx))D^T\bz f^T(\bx)\right],\nwl
&W_{k,f}=\bbE\left[\Gamma(f(\bx))D^T\bz \left(f^k\right)^T(\bx)\right],\ k=1,2,\ldots,\nwl
&b_f=\bbE\left[\Gamma(f(\bx))D^T\bz\right]
\end{eqnarray}
are specialized values of $W_0,W_k,b$, respectively.
\subsection{Initialization}
An efficient execution of the above procedure requires us to judiciously select the parameters $U,D,c$. We select $U$ as the collection of basis vectors of the first $d$ principal components of $\bx$, i.e. $U=P_1^T$ where $\bbE[\bx\bx^T]=P\Sigma P^T$ is the Eigen-representation (SVD) of the correlation matrix, $P=[p_1\ p_2\ \ldots p_n]$ and $P_1=[p_1\ p_2\ \ldots p_d]$. The matrices $U,c$ are selected by minimizing the MSE objective with $f=f_0$. This yields,
\[D=\bbE\left[\by f^T_0(\bx)\right]\bbE\left[f_0(\bx) f^T_0(\bx)\right]^{-1},\]
\[c=\bbE\left[\by\right]-D\bbE\left[f_0(\bx)\right].\]
This also affords us to update these matrices in the course of the optimization,
\[D\gets\bbE\left[\by f^T_t(\bx)\right]\bbE\left[f_t(\bx) f^T_t(\bx)\right]^{-1},\]
\[c\gets\bbE\left[\by\right]-D\bbE\left[f_t(\bx)\right].\]
\subsection{Momentum Method}
Momentum methods are popular in machine learning and lead to considerable improvement in both performance and convergence speed   \cite{sutskever2013importance, kingma2014adam}. Since the originally formulated do not conform to our  geometric regularization framework, we proposed an alternative approach to effectively mix the learning direction $\phi_f$ at each iteration with its preceding iterates to better control any associated rapid changes over iterations (low-pass filtering). Here to keep the geometric regularization structure, we instead mix the parameters $W_k,b$. This leads to the following modification in the original algorithm in \eqref{eq:step_simple}:
\begin{eqnarray}\label{eq:momentum}
&f_{t+1}=f_t+\mu_t\Gamma(f)\left[V_{0,t}f_t+\suml_{k=1}^rV_{k,t}f^k+e_t\right],
\nwl
&V_{k,t+1}=\alpha_k V_{k,t}+ W_{k,f_t},\ k=0,1,\ldots,r,
\nwl
&e_{t+1}=\beta e_{t}+ b_{f_t},
\end{eqnarray}
where $W_{k,f_t}, b_{f_t}$ are given in \eqref{eq:params}. 
\subsection{Learning Parameter Selection}
To select  the remaining parameters $\alpha_k,\beta$ and $\mu_t$, we manually tune parameters $\alpha_k,\beta$, specifically  utilize two strategies when tuning $\mu_t$: a) fixing $\mu_t=\mu$ and b) using line search. The second method, we obtain by simple computations as
\[
\mu_t=\frac{\bbE[\bz_t^TD\psi_t]}{\bbE[\|D\psi_t\|_2^2]},
\]
where $\bz_t=\by-Df_t(\bx)-c$, and 
\[
\psi_t=\Gamma(f_t)\left[V_{0,t}f_t+\suml_{k=1}^rV_{k,t}f^k+e_t\right].
\]
\subsection{Incorporating Shift Invariance}
In the context of deep learning, especially for image processing, convolutional networks are popular. They differ from the regular deep networks in attempting to induce shift invariance in the  linear operations of some layers, by way  of  convolution (Toeplitz matrix). We may adopt the same strategy in geometric regularization by further assuming that $W_0$ in \eqref{eq:GR_general} represents a convolution. We skip the derivations, for not only space limitation reasons, but also for their similarity to those leading to \eqref{eq:params}. The resulting algorithm with the momentum method is similar to \eqref{eq:momentum} where $W_{0,f}$ is replaced by $W_{\mathrm{conv},f}$, defined as
\[
W_{\mathrm{conv},f}=\arg\maxl_W\langle W, W_{0,f}\rangle,
\] 
where the optimization is over unit-norm convolution (Toeplitz) matrices. It turns out that since Toeplitz matrices form a vector space,  $W_{\mathrm{conv},f}$ is a linear function of $W_{0,f}$ and can be quickly calculated \cite{bottcher2012toeplitz}. Due to space limitation, we defer the details to \cite{future_paper}. 

\section{Theoretical Discussion}
\subsection{Relation to Deep Residual Networks}
\begin{figure}[t]
\centering
\includegraphics[width=7cm, height=4cm]{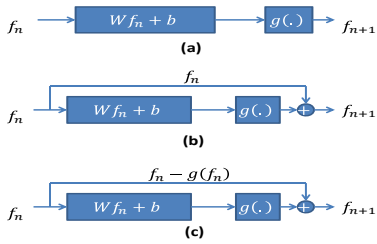}
\caption{Schematic scheme of a single layer in a)ANN  b)ResNet and c)modified ResNet Architectures.}
\label{fig:ResNet}
\end{figure}

The proposed geometric regularization for nonlinear regression in \eqref{eq:GR_general} is inspired by the advances in the field of neural networks and deep learning. Recall that a generic deep artificial neural network (DNN) represents a sequence of functions (hidden layers) $f_0(x), f_1(x),\ldots, f_T(x)$ where $f_0(x)=x$ and $f_t$ for $t=0,1,2,\ldots$, is $d_t$-dimensional, where $d_t$ is the network width in the $t^\tth$ layer. The relation of these functions is plotted in Figure \ref{fig:ResNet} (a). 
The so-called residual network (ResNet) architecture modifies  DNNs by introducing bridging branches (edges) as shown by Figure \ref{fig:ResNet} (b).
We observe that the geometric regularization in \eqref{eq:GR_general}  corresponds to a modified version of ResNets, as depicted in Figure \ref{fig:ResNet} (c), which can be written as
\begin{equation}\label{eq:mod_resnet}
f_{t+1}=g(W_tf_t+b_t)-g(f_t)+f_{t}.
\end{equation}
More concretely, when $W_t, b_t$ are respectively near identity and near zero, i.e. $W_t=I+\epsilon \barW_t$ and $b_t=\epsilon\barb_t$ for small values of $\epsilon$, we observe by Taylor expansion of \eqref{eq:mod_resnet} with respect to $\epsilon$ that the differential equation in \eqref{eq:integral} with geometric regularization in \eqref{eq:GR_general} provides the limit of the above modified ResNet architecture. This profound relation provides a novel approach for analyzing deep networks, which is deferred to \cite{future_paper}.

\section{Numerical Results}
\label{sec:results}
As a preliminary validation, we examine geometric regularization on the MNIST handwritten digits database including $50,000$ $28\times 28$ black and white images of handwritten digits for training and $10,000$ more for testing \cite{lecun1998gradient,ciregan2012multi}. We note that state-of-the-art techniques already achieve an accuracy as high as $99.7\%$, thus justifying our validation study merely as a proof of concept.  We use a single fixed function $f^1(x)=x$. In all experiments, we set $\alpha_0=\beta=0.98$ and let $\alpha_1$ vary. 
\begin{figure}
\centering
\begin{tabular}{|c|c|}
\hline
Method & Performance \\
\hline
plain iterations  & 97.4\%\\
convolutional iterations & 98.0\%\\
2-stage learning & 98.7\%\\
\hline
\end{tabular}
\caption{Performance of different learning strategies}
\label{table:1}
\end{figure}

We have performed extensive numerical studies with different strategies (fixed or variable $D,c$, different step size selection methods and convolutional/plain layers), but can only focus on some key results due to space limitation. A more comprehensive comparison between these strategies is also insightful \cite{future_paper}. 
A summary of the best achieved performances is listed in Figure \ref{table:1}, where plain (non-convolutional) iterations are applied with fixed step size $\mu=0.06$ and $\alpha_1=0.99$, $d=400$ and fixed $D,c$. The convolutional iterations also include 2-D convolutional (Toeplitz) matrices with window length 5, fixed step size $\mu=1$ and $\alpha_1=0.95$, and $D,c$ updated at each iteration. We also consider a 2-stage procedure, where in the first $50$ iterations, convolutional matrices are considered, and plain iterations are subsequently applied. In both stages, the step size is fixed to $\mu=3$ and $\alpha=0.95$, while the matrices $D,c$ are updated at each iteration.

\begin{figure}[t]
\centering
\includegraphics[width=8cm,height=4.8cm]{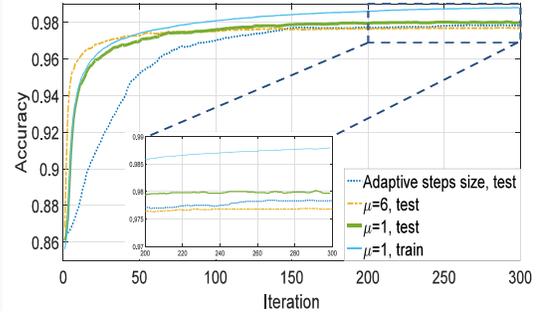}
\caption{Performance of different step size selection strategies.}
\label{fig:perform}
\end{figure}
Figure \ref{fig:perform} compares different strategies for step size selection with convolutional iterations by their associated performance, i.e. the fraction of correctly classified images in different iterations. The best asymptotic performance is obtained by fixing $\mu=1$. Faster convergence may be obtained by larger step sizes at the expense of a decreased asymptotic performance. For example for $\mu=6$, the algorithms reaches $96\%$ accuracy in only $10$ iterations and is $97\%$ correct at $30$. However, the process becomes substantially slower afterwards, which suggests a multi-stage procedure to boost performance. Using adaptive step size with line search shows a slightly degraded (higher than $\mu=6$) performance, but dramatically decreases the convergence rate.

\section{Conclusion}
We proposed a supervised learning technique, which enjoys many common properties with deep learning, such as successive application of linear and non-linear operators, momentum method of implementation and convolutional layers. In contrast to deep learning, our method abandons the need for back propagation to hence improve the computational burden. Our method is semi-parametric as it essentially exploits a large number of weight parameters, yet avoiding over-parametrization. Another advantage of our technique is that it can theoretically be analyzed by tools in differential geometry as briefly discussed earlier. A conprehensive development is in \cite{future_paper}. The performance on the data sets we have thus far achieved, promises a great and unexplored  potential waiting to be unveiled.



%
%
%
%


\bibliographystyle{IEEEtran}
\bibliography{refs}

\end{document}

%% file: fonts.tex
\newcommand{\bGamma}{\bm{\Gamma}}
\newcommand{\bSigma}{\bm{\Sigma}}
\newcommand{\bOmega}{\bm{\Omega}}

\newcommand{\bdelta}{\bm{\delta}}
\newcommand{\bomega}{\bm{\omega}}
\newcommand{\bgamma}{\bm{\gamma}}
\newcommand{\bepsilon}{\bm{\epsilon}}
\newcommand{\blambda}{\bm{\lambda}}
\newcommand{\btheta}{\bm{\theta}}
\newcommand{\bpsi}{\bm{\psi}}
\newcommand{\bmeta}{\bm{\eta}}
\newcommand{\bzeta}{\bm{\zeta}}
\newcommand{\bmu}{\bm{\mu}}
\newcommand{\bnu}{\bm{\nu}}
\newcommand{\bpi}{\bm{\pi}}
\newcommand{\bsigma}{\bm{\sigma}}

\newcommand{\tilDelta}{\tilde{\Delta}}
\newcommand{\tlDelta}{\tilde{\Delta}}
\newcommand{\tlepsilon}{\tilde{\epsilon}}
\newcommand{\tltheta}{\tilde{\theta}}

\newcommand{\bA}{\mathbf{A}}
\newcommand{\bE}{\mathbf{E}}

\newcommand{\bG}{\mathbf{G}}
\newcommand{\bH}{\mathbf{H}}
\newcommand{\bI}{\mathbf{I}}
\newcommand{\bJ}{\mathbf{J}}
\newcommand{\bL}{\mathbf{L}}
\newcommand{\bM}{\mathbf{M}}
\newcommand{\bN}{\mathbf{N}}
\newcommand{\bP}{\mathbf{P}}
\newcommand{\bQ}{\mathbf{Q}}
\newcommand{\bR}{\mathbf{R}}
\newcommand{\bS}{\mathbf{S}}
\newcommand{\bT}{\mathbf{T}}
\newcommand{\bW}{\mathbf{W}}
\newcommand{\bX}{\mathbf{X}}
\newcommand{\bY}{\mathbf{Y}}
\newcommand{\bZ}{\mathbf{Z}}

\newcommand{\ba}{\mathbf{a}}
\newcommand{\bb}{\mathbf{b}}
\newcommand{\bd}{\mathbf{d}}
\newcommand{\be}{\mathbf{e}}
\newcommand{\mbf}{\mathbf{f}}
\newcommand{\bg}{\mathbf{g}}
\newcommand{\bh}{\mathbf{h}}
\newcommand{\bl}{\mathbf{l}}
\newcommand{\bn}{\mathbf{n}}
\newcommand{\bp}{\mathbf{p}}
\newcommand{\bq}{\mathbf{q}}
\newcommand{\br}{\mathbf{r}}
\newcommand{\bs}{\mathbf{s}}
\newcommand{\bu}{\mathbf{u}}
\newcommand{\bv}{\mathbf{v}}
\newcommand{\bw}{\mathbf{w}}
\newcommand{\bx}{\mathbf{x}}
\newcommand{\by}{\mathbf{y}}
\newcommand{\bz}{\mathbf{z}}

\newcommand{\hbeta}{\hat{\beta}}
\newcommand{\htheta}{\hat{\theta}}
\newcommand{\hsigma}{\hat{\sigma}}

\newcommand{\hp}{\hat{p}}
\newcommand{\hr}{\hat{r}}
\newcommand{\hs}{\hat{s}}
\newcommand{\hx}{\hat{x}}

\newcommand{\hN}{\hat{N}}

\newcommand{\hbSigma}{\hat{\bm{\Sigma}}}

\newcommand{\hba}{\hat{\mathbf{a}}}
\newcommand{\hbs}{\hat{\mathbf{s}}}
\newcommand{\hbx}{\hat{\mathbf{x}}}
\newcommand{\hbv}{\hat{\mathbf{v}}}

\newcommand{\hbW}{\hat{\mathbf{W}}}

\newcommand{\dif}{\text{d}}

\newcommand{\bbC}{\mathbb{C}}
\newcommand{\bbE}{\mathbb{E}}
\newcommand{\bbR}{\mathbb{R}}
\newcommand{\bbN}{\mathbb{N}}
\newcommand{\bbZ}{\mathbb{Z}}

\newcommand{\calA}{\mathcal{A}}
\newcommand{\calB}{\mathcal{B}}
\newcommand{\calC}{\mathcal{C}}
\newcommand{\calD}{\mathcal{D}}
\newcommand{\calE}{\mathcal{E}}
\newcommand{\calF}{\mathcal{F}}
\newcommand{\calH}{\mathcal{H}}
\newcommand{\calL}{\mathcal{L}}
\newcommand{\calN}{\mathcal{N}}
\newcommand{\calM}{\mathcal{M}}
\newcommand{\calP}{\mathcal{P}}
\newcommand{\calS}{\mathcal{S}}
\newcommand{\calT}{\mathcal{T}}
\newcommand{\calV}{\mathcal{V}}
\newcommand{\calW}{\mathcal{W}}
\newcommand{\calX}{\mathcal{X}}
\newcommand{\calY}{\mathcal{Y}}

\newcommand{\calhL}{\mathcal{\hat{L}}}

\newcommand{\tlA}{\tilde{A}}
\newcommand{\tlC}{\tilde{C}}

\newcommand{\tlv}{\tilde{v}}
\newcommand{\tls}{\tilde{s}}

\newcommand{\barb}{\bar{b}}
\newcommand{\barm}{\bar{m}}
\newcommand{\barn}{\bar{n}}
\newcommand{\barr}{\bar{r}}
\newcommand{\bary}{\bar{y}}

\newcommand{\barC}{\bar{C}}
\newcommand{\barH}{\bar{H}}
\newcommand{\barK}{\bar{K}}
\newcommand{\barL}{\bar{L}}
\newcommand{\barW}{\bar{W}}

\newcommand{\barba}{\bar{\ba}}
\newcommand{\barby}{\bar{\by}}
\newcommand{\barbz}{\bar{\bz}}

\newcommand{\tlbA}{\tilde{\bA}}
\newcommand{\tlbE}{\tilde{\bE}}
\newcommand{\tlbW}{\tilde{\bW}}

\newcommand{\tlbv}{\tilde{\bv}}

\newcommand{\tc}{\text{c}}
\newcommand{\td}{{\text{d}}}

\newcommand{\bzero}{\mathbf{0}}

\newcommand{\suml}{\sum\limits}
\newcommand{\minl}{\min\limits}
\newcommand{\maxl}{\max\limits}
\newcommand{\infl}{\inf\limits}
\newcommand{\supl}{\sup\limits}
\newcommand{\liml}{\lim\limits}
\newcommand{\intl}{\int\limits}
\newcommand{\bigcupl}{\bigcup\limits}

\newcommand{\opconv}{\text{conv}}

\newcommand{\eref}[1]{(\ref{#1})}

\newcommand{\sinc}{\text{sinc}}
\newcommand{\tr}{\text{Tr}}
\newcommand{\var}{\text{Var}}
\newcommand{\cov}{\text{Cov}}
\newcommand{\tth}{\text{th}}

\newcommand{\nwl}{\nonumber\\}

\newenvironment{vect}{\left[\begin{array}{c}}{\end{array}\right]}
\newtheorem{theorem}{Theorem}
\newtheorem{lemma}{Lemma}